\documentclass[journal]{IEEEtran}
\IEEEoverridecommandlockouts
\usepackage{cite}
\usepackage{amsmath,amssymb,amsfonts}
\usepackage{graphicx}
\usepackage{textcomp}
\usepackage{xcolor}

\usepackage{algorithm}
\usepackage{algpseudocode}
\usepackage[percent]{overpic}
\usepackage{multirow}
\usepackage{url}
\usepackage[bookmarksopen,bookmarksdepth=2,breaklinks=true]{hyperref}

\def\R{\mathbb R}

\def\BibTeX{{\rm B\kern-.05em{\sc i\kern-.025em b}\kern-.08em
    T\kern-.1667em\lower.7ex\hbox{E}\kern-.125emX}}
\begin{document}

\title{Deep Sparse Representation-based Classification}

\author{{Mahdi~Abavisani,~\IEEEmembership{Student Member,~IEEE} and
Vishal M.		Patel,~\IEEEmembership{Senior Member,~IEEE}} 
\thanks{M.		Abavisani is with the department of Electrical and Computer Engineering at Rutgers University, 
   Piscataway, NJ USA.  email: mahdi.abavisani@rutgers.edu.		Vishal M.	Patel is with the department of Electrical and Computer Engineering at Johns Hopkins University, Baltimore, MD USA.		email: vpatel36@jhu.edu.   This work was partially supported by the NSF grant 1618677 and by US Office of Naval Research (ONR) Grant YIP N00014-16-1-3134.}
}

\maketitle

\begin{abstract}
We present a transductive deep learning-based formulation for the sparse representation-based classification (SRC) method.   The proposed network consists of a convolutional autoencoder along with a fully-connected layer.  The role of the autoencoder network is to learn robust deep features for classification.  On the other hand, the fully-connected layer, which is placed in between the encoder and the decoder networks, is responsible for finding the sparse representation.  The estimated sparse codes are then used for classification. Various experiments on three different datasets show that the proposed network leads to sparse representations that give better classification results than state-of-the-art SRC methods.   The source code is available at: ~\href{https://github.com/mahdiabavisani/DSRC}{github.com/mahdiabavisani/DSRC}.
\end{abstract}
\begin{IEEEkeywords}
Deep learning, sparse representation-based classification, deep sparse representation-based classification.
\end{IEEEkeywords}
\section{Introduction}
\label{sec:intro}
Sparse coding has become widely recognized as a powerful tool in signal processing and machine learning with various applications in computer vision and pattern recognition \cite{SRC_PAMI_2009, aharon2006k, SSC_PAMI}.  Sparse representation-based classification (SRC) as an application of sparse coding was first proposed in \cite{SRC_PAMI_2009}, and was shown to provide robust performance on various face recognition datasets.  Since then, SRC has been used in numerous applications \cite{mei2011robust,shekhar2014joint,abavisani2015robust,joneidi2015lfm,perera2019face,ghassemi2019learning}.   In SRC, an unlabeled sample is represented as a sparse linear combination of the labeled training samples.  This representation is obtained by solving a sparsity-promoting optimization problem.   Once the representation is found, the label is assigned to the test sample based on the minimum reconstruction error rule \cite{SRC_PAMI_2009}.  

The SRC method is based on finding a linear representation
of the data. However, linear representations are almost always inadequate for representing non-linear structures of the data which arise in many practical applications. To deal with this issue, some works have exploited the kernel trick to develop  non-linear extensions of the SRC-based methods \cite{SRC_MKL,KKSVD, KSSC, SSC_ICCV2013,KSRC2012,Kernel_SR1,Kernel_SR2,Kernel_CS,van2012kernel,yin2012kernel,latenst_SSC_LRR}.  
  Kernel SRC methods require the use of a pre-determined kernel function such as polynomial or Gaussian. Selection of the kernel function and its
parameters is an important issue in training when kernel SRC
methods are used for classification.

In this paper, we propose a deep neural network-based framework that finds an explicit nonlinear mapping of data, while simultaneously obtaining sparse codes that can be used for classification.      Learning nonlinear mappings with neural networks has been shown to produce remarkable improvements in subspace clustering tasks~\cite{deepsc17nips,abavisani2018deep}.  We introduce a transductive model, which accepts a set of training and test samples, learns a mapping that is suitable for sparse representation, and recovers the corresponding sparse codes.    Our model consists of an encoder that is responsible for learning the mapping, a sparse coding layer which mimics the task of constructing the mapped test samples by a combination of the mapped training samples, and a decoder that is used for training the networks.

\begin{figure}[t!]
  \centerline{\includegraphics[width=0.5\textwidth]{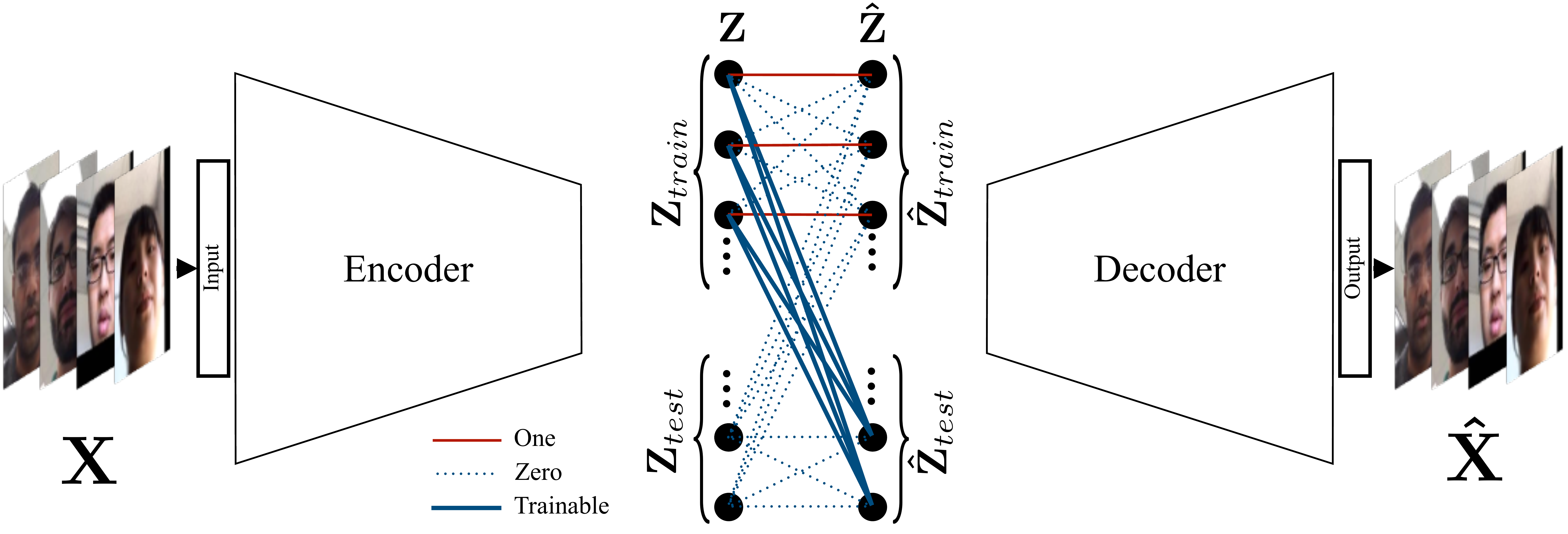}}\vskip -5pt
\caption{An overview of the proposed deep SRC network.  The trainable parameters of sparse coding layer are depicted with solid blue lines. Note that $\mathbf{Z}_{train} = \mathbf{\hat Z}_{train}$, and $\mathbf{Z}_{test} \approx \mathbf{\hat Z}_{test}= \mathbf{Z}_{train}\mathbf{A}$.}
\label{fig:diagram}
\end{figure} 
\vskip -5pt

\subsection{Sparse representation-based classification}\label{sec:SRC}
In SRC, given a set of labeled training samples, the goal is to classify an unseen set of test samples.  Suppose that we collect all the vectorized training samples with the label $i$ in the matrix $\mathbf{X}^{i}_{train}\in \R^{d_0\times{n_i}}$, where $d_0$ is the dimension of each sample and $n_i$ is the number of samples in class $i$, then the training matrix can be constructed as
\begin{equation}\label{eq:x}
   \mathbf{X}_{train} =  [\mathbf{X}^{1}_{train},\mathbf{X}^{2}_{train},\cdots, \mathbf{X}^{K}_{train}]  \in \R^{d_0\times n}
\end{equation}
where $n_1+n_2+\cdots+n_K=n$ and we have a total of $K$ classes.

In SRC, it is assumed that an observed sample $\mathbf{x}_{test} \in \R^{d_0}$ can be well approximated by a linear combination of the samples in $\mathbf{X}^{i}_{train}$ if $\mathbf{x}_{test}$ is from class $i$.    Thus, it is possible to predict the class of a given unlabeled data by finding a set of samples in the training set that can better approximate $\mathbf{x}_{test}$.  Mathematically, these samples can be found by solving the following optimization problem
\begin{equation}\label{eq:x0}
 \min_{\boldsymbol{\alpha}} \|\boldsymbol{\alpha} \|_{0} \;\;\text{s.t.}\;\; \mathbf{x}_{test} =\mathbf{X}_{train}\boldsymbol{\alpha}, 
 \end{equation}
 where $\|\boldsymbol{\alpha}\|_0$ counts the number of non-zero elements in $\boldsymbol{\alpha}$. The minimization problem \eqref{eq:x0} finds a sparse solution for the linear system. However, since the optimization problem \eqref{eq:x0} is an NP-hard problem, in practice, a sparsity constraint is enforced by the $\ell_1$-norm of $\boldsymbol{\alpha}$ which is a convex relaxation of the above problem ~\cite{candes2005decoding,donoho2006most}.    Thus, in practice the following minimization problem is solved to obtain the sparse codes 
\begin{equation}\label{eq:x1}
  \min_{\boldsymbol{\alpha}} \| \mathbf{x}_{test}  - \mathbf{X}_{train}\boldsymbol{\alpha} \|^2_2 + \lambda_0 \|\boldsymbol{\alpha} \|_{1}, 
 \end{equation} 
 where $\lambda_0$ is a positive regularization parameter.  Once $\boldsymbol{\alpha}$ is found, one can estimate the class label of $\mathbf{x}_{test}$ as follows
\begin{equation}\label{eq:classify0}
 \text{class}(\mathbf{x}_{test}) = \text{arg}\min_{k} \|\mathbf{x}_{test} - \mathbf{X}_{train}\delta_{k}(\boldsymbol{\alpha}) \|^2_{2},
 \end{equation}
where $\delta_{k}(\cdot)$  is the characteristic function that selects the coefficients associated with the class $i$.

\section{Deep sparse representation-based classification network}
\label{sec:proposed}
We develop a transductive classification model based on sparse representations. In a transductive model, as opposed to inductive models, both training and test sets are observed, and the learning process pursues reasoning from the specific training samples to a specific set of test cases~\cite{gammerman1998learning}.  We build our method based on convolutional autoencoders.   In particular, our network contains an encoder, a sparse coding layer, and a decoder.   The encoder receives both the training and test sets as raw data inputs and extracts abstract features from them.  The sparse coding layer recovers the test cases by a sparse linear combination of the training samples, and concatenates them along with the training features which are then fed to the  decoder.  The decoder maps both the training embeddings and the recovered test embeddings back to the original representation of the data.  Figure~\ref{fig:diagram} gives an overview of the proposed deep SRC (DSRC) framework.\\

\noindent \textbf{Sparse representation: }
Let $\mathbf{X}_{train} \in \R^{d_0 \times n}$ and, $\mathbf{X}_{test} \in \R^{d_0 \times m}$ be the given vectorized training and testing data, respectively.   We feed $\mathbf{X} = [\mathbf{X}_{train},\mathbf{X}_{test}] $ to the encoder, where it develops the corresponding embedding features $\mathbf{Z} = [\mathbf{Z}_{train},\mathbf{Z}_{test}] \in \R^{d_z \times (m+n)} $.  The minimization problem \eqref{eq:x1} for a single test observation can be re-written for a matrix of testing embedding features as
\begin{equation}\label{eq:z1}
  \min_{\mathbf{A}} \| \mathbf{Z}_{test}  - \mathbf{Z}_{train}\mathbf{A} \|^2_F + \lambda_0 \|\mathbf{A} \|_{1}, 
 \end{equation}
where $\mathbf{A} \in R^{n \times m}$ is the coefficient matrix that contains the sparse codes in its columns, and $\lambda_0$ is a positive regularization parameter.  Note that the first penalty term in  equation~\eqref{eq:z1} is equivalent to the penalty term used for a fully-connected neural network layer with the input of $\mathbf{Z}_{train}$, the output of  $\mathbf{Z}_{test}$ and trainable parameters of $\mathbf{A}$.  As a result, considering the sparsity constraint, one can model the optimization problem~\eqref{eq:z1} in a neural network framework with a fully-connected layer with sparse parameters which have no non-linearity activation or bias nodes.  We use such a model inside our sparse coding layer to find the sparse codes for the observed test set.

The sparse coding layer is located between the encoder and decoder networks. Its task for $\mathbf{Z}_{train}$ is to pass them to the decoder, and for the test features $\mathbf{Z}_{test}$ it will pass their reconstructions that are found from~\eqref{eq:z1}, as $\mathbf{Z}_{train}\mathbf{A}$, to the decoder. 
Thus, assuming that $\mathbf{\hat Z}_{train}$ and $\mathbf{\hat Z}_{test}$ are the outputs of the sparse coding layer for training and testing features, we have 
 \begin{equation}\label{eq:zs}
\mathbf{\hat Z}_{train} = \mathbf{Z}_{train}\mathbf{I}_{n}, \;\;\;
\mathbf{\hat Z}_{test} = \mathbf{Z}_{train}\mathbf{A},
 \end{equation}
 where $\mathbf{I}_{n} \in \R^{n\times n}$ is the identity matrix. Therefore, if the decoder's input is $\mathbf{\hat Z} = [\mathbf{\hat Z}_{train},\mathbf{\hat Z}_{test}]$, from \eqref{eq:zs} we can calculate  $\mathbf{\hat Z}$ as $\mathbf{\hat Z} = \mathbf{Z}\boldsymbol{\Theta}_{sc}$,
where 
 \begin{equation}\label{eq:theta}
\boldsymbol{\Theta}_{sc}  =   \left[ \begin{gathered}
  {{\mathbf{I}}_{n}}\;\;\; \hfill  \mathbf{A} \hfill\\
  {\mathbf{0}_{n\times m}}\; {\mathbf{0}_{m}} \hfill \\ 
\end{gathered}  \right]. 
 \end{equation}
In equation~\eqref{eq:theta}, $\mathbf{0}_{n\times m} \in \R^{n \times m}$ and $\mathbf{0}_{m} \in \R^{m \times m}$ are zero matrices. One can write an end-to-end training objective that includes sparse coding and training of the encoder-decoder as follows
 \begin{equation}\label{eq:loss}
  \min_{\boldsymbol{\Theta}} \| \mathbf{Z}  - \mathbf{Z}\boldsymbol{\Theta}_{sc} \|^2_F + \lambda_0 \|\boldsymbol{\Theta}_{sc} \|_{1} + \lambda_1 \| \mathbf{X}  - \mathbf{\hat X}\|^2_F, 
 \end{equation}
 where $\boldsymbol{\Theta}$ is the union of all the trainable parameters including encoder and decoder's parameters and $\mathbf{A}$. Here, $\mathbf{\hat X} = [\mathbf{\hat X}_{train},\mathbf{\hat X}_{test}]$ is the output of the decoder (i.e. reconstructions), and  $\lambda_0$ and $\lambda_1$ are positive regularization parameters. Note that the optimization problem~\eqref{eq:loss} simultaneously finds sparse codes $\mathbf{A}$ and a set of desirable embedding features $\mathbf{Z}$ that are especially suitable for providing efficient sparse codes.\\

 \noindent \textbf{Classification:}
 Once the sparse coefficient matrix $\mathbf{A}$ is found, it can be used for associating the class labels to the test samples.  For each test sample $\mathbf{x}^{i}_{test}$ in $\mathbf{X}_{test}$, its embedding features $\mathbf{z}^{i}_{test}$, and the corresponding sparse code column $\boldsymbol{\alpha}^{i}$ in $\mathbf{A}$ are used to estimate the class labels as follows 
 \begin{equation}\label{eq:classify}
 \text{class}(\mathbf{x}^{i}_{test}) = \text{arg}\min_{k} \|\mathbf{z}^{i}_{test} - \mathbf{Z}_{train}\delta_{k}(\boldsymbol{\alpha}^{i}) \|^2_{2}.
 \end{equation}
 The proposed DSRC method is summarized in Algorithm~\ref{alg:dsrc}.

\begin{algorithm}[t]
\caption{Deep sparse representation-based classification}\label{alg:dsrc}
\begin{algorithmic}[1]
\Procedure{{DSRC}}{$\mathbf{X}_{train}, \mathbf{X}_{test}, \lambda_0, \lambda_1$}.
\State Construct $\mathbf{X} = [\mathbf{X}_{train}, \mathbf{X}_{test}]$.
\State Find $\mathbf{A}$ via $\boldsymbol{\Theta}$ by solving the optimization problem~\eqref{eq:loss}.
\State Classify the test samples using \eqref{eq:classify} .
\EndProcedure 
\end{algorithmic}
\end{algorithm}

\section{Experimental results}
\label{sec:results}
\begin{figure}[t]
 \centerline{\begin{overpic}[width=0.3\textwidth,tics=3]{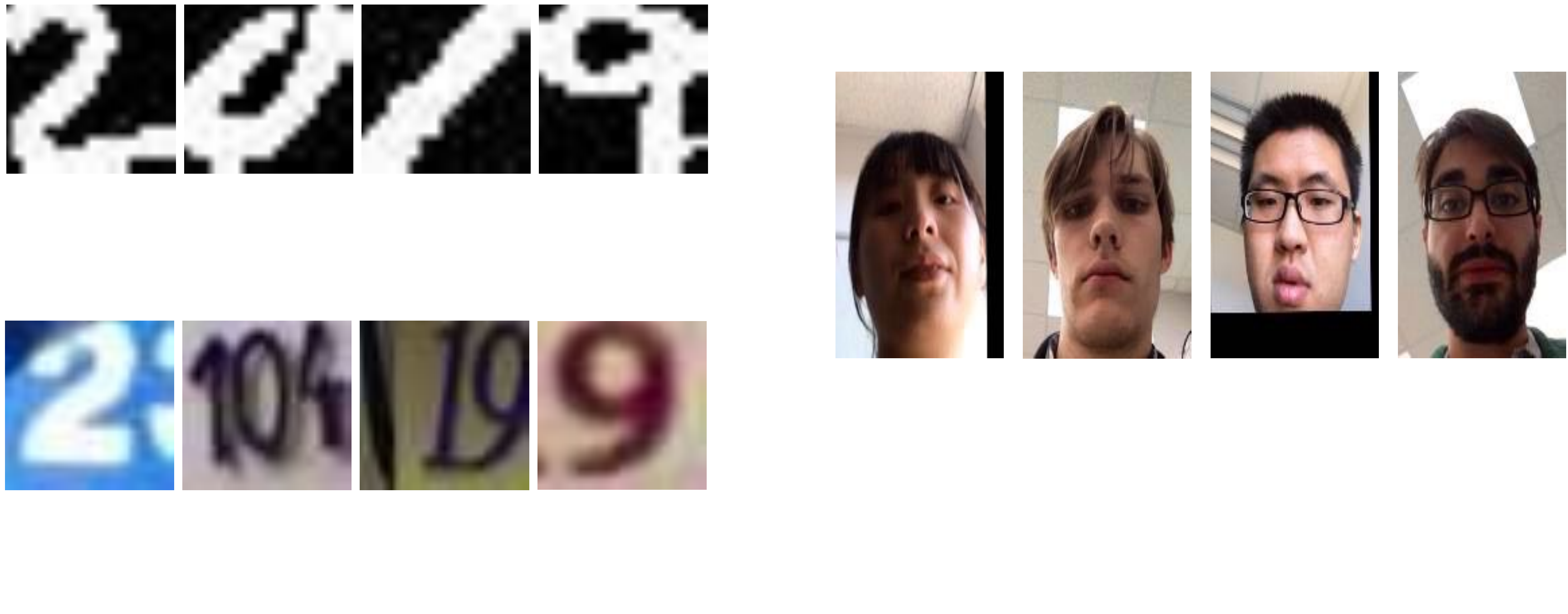}
 \put (11,24) {\tiny{(a) USPS~\cite{hull1994database}}}
  \put (11,3) {\tiny{(b) SVHN~\cite{netzer2011reading}}}
 \put (59,12) {\tiny{(c) UMDAA-01~\cite{zhang2015domain}}}
 \end{overpic}}
\vskip -12pt\caption{Sample images from (a) USPS~\cite{hull1994database}, (b) SVHN~\cite{netzer2011reading}, and (c) UMDAA-01~\cite{zhang2015domain}.}
\label{fig:datasets}
\end{figure}

In this section, we evaluate our method against state-of-the-art SRC methods.  The USPS handwritten digits dataset~\cite{hull1994database}, the street view house numbers (SVHN) dataset~\cite{netzer2011reading}, and the UMDAA-01 face recognition dataset~\cite{zhang2015domain} are used in our experiments.  Figure~\ref{fig:datasets} (a), (b), and (c) show sample images from these datasets.  Since the number of parameters in the sparse coding layer scales with the multiplication of training and testing sizes, we randomly select a smaller subset of the used datasets and perform all the experiments on the selected subset.  In all the experiments, the input images are resized to  $32 \times 32$. 
 
\begin{figure}[t]
  \centerline{\includegraphics[width=0.50\textwidth]{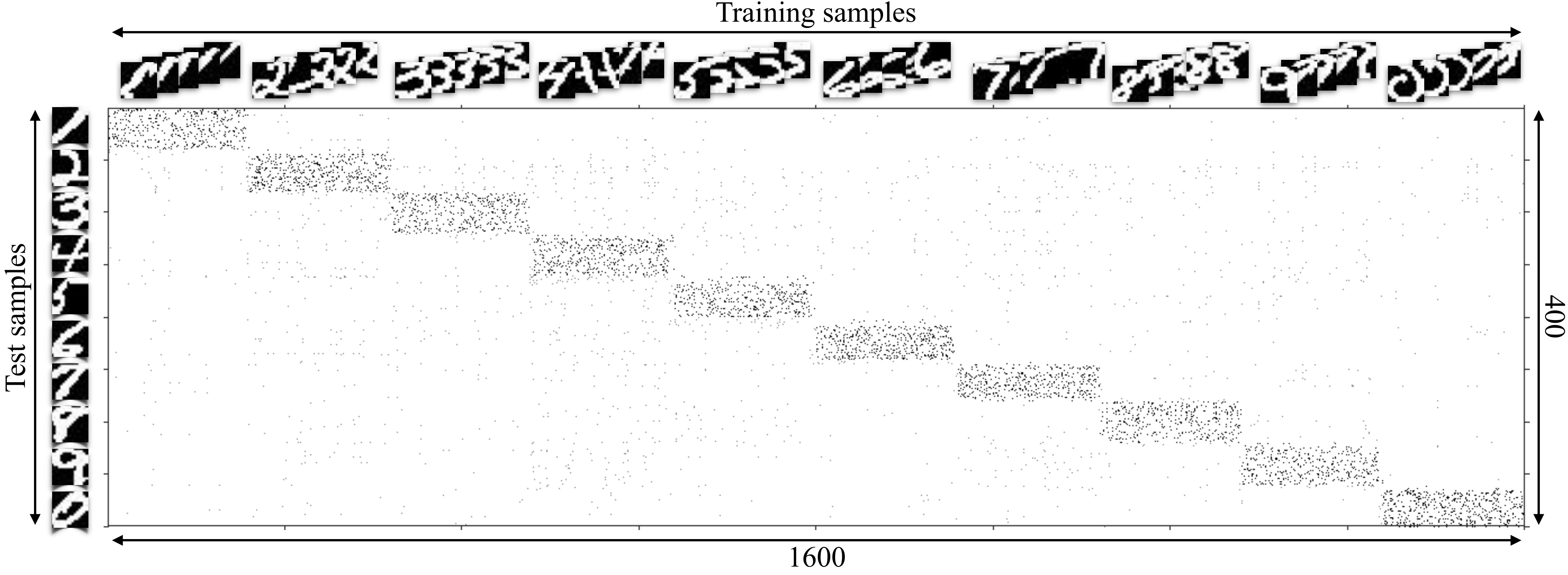}}
\caption{Visualization of the sparse coding matrix ($\mathbf{A}$) in the experiment with the USPS dataset. Note that for better visualization the absolute value of the transposed $\mathbf{A}$ (i.e. $|\mathbf{A}^T|$) is shown. }
\label{fig:res}
\end{figure}

We compare our method with the standard SRC method~\cite{SRC_PAMI_2009}, Kernel SRC  (KSRC)~\cite{KSRC2012}, SRC on features extracted from an autoencoder with similar architecture to our network (AE-SRC), and SRC on features extracted from the state-of-the-art pre-trained networks.  In our experiment with the pre-trained networks, the networks are pre-trained on the Imagenet dataset~\cite{deng2009imagenet}.   For this purpose, we use the following four popular network architectures: VGG-19\cite{simonyan2014very},  Inception-V3~\cite{szegedy2016rethinking}, Resnet-50~\cite{he2016deep} and Densenet-169~\cite{huang2017densely}.   We feed these networks with our datasets, extract the features corresponding to the last layer before classification, and pass them to the classical SRC algorithm.  Note these networks accept $224\times224$ inputs. Thus, as a preprocessing step, we resample the input images to  $224\times224$ images before feeding them to the pre-trained networks.

We compare different methods in terms of their five-fold averaged classification accuracy.  In all the experiments, unless otherwise stated, we randomly split the datasets into sets of training and testing samples, where $20\%$ of the samples are used for testing, and $80\%$ of the samples are used as the training set.

\noindent \textbf{Network structure:}   The encoder network of our model  consists of stacked four convolutional layers, and the decoder has three fractionally-strided convolution layers (also known as deconvolution layers).  Each plugged in convolution or fractionally-strided convolution is coupled with a ReLu nonlinearity as well, but does not have a batch-norm layer. Table~\ref{tbl:nets} gives the details of the network, including the  kernel sizes and the number of filters.

\begin{table}[htp!]
	\centering
	\caption{Details of our networks. Note that the number of parameters in the sparse coding layer rely on the size of dataset including the $n$ training and $m$ test samples. }
	\label{tbl:nets}
	\resizebox{\linewidth}{!}{
		\begin{tabular}{|c|c|c|c|l|  p{0.7cm}|}
			\cline{2-6}
			\multicolumn{1}{c|}{ }  &  \multirow{3}{*}{Layer}  &  \multirow{3}{*}{Input}  &  \multirow{3}{*}{Output}  &  \multirow{3}{*}{Kernel}  &  (stride, pad)  \\
			\cline{2-6}\hline	
			\multirow{4}{*}{Encoder}   &  Conv  1  &   $\mathbf{X}$   &  Conv 1   &   $1\times5\times5\times10$  &  (2,1) \\
			 &     Conv 2  &   Conv 1   &  Conv 2   & $1\times3\times3\times20$  &  (2,1) \\			
			 &     Conv 3  &   Conv 2   &  Conv 3   &   $1\times3\times3\times30$  &  (1,0) \\	
			 &     Conv 4  &   Conv 3   & $\mathbf{Z}$     &   $1\times3\times3\times30$  &  (1,0) \\
			\hline\hline	
			\multirow{1}{*}{Sparse coding layer}   &  \multirow{1}{*}{ $\boldsymbol{\Theta}_{sc}$}  &   $\mathbf{Z}$     &  $\mathbf{\hat Z}$  &  \centering \multirow{1}{*}{$m\times n$ Parameters}  &  \multirow{1}{*}{-} \\
			\hline\hline		
			\multirow{3}{*}{Decoder}   &  deconv 1  &  $\mathbf{\hat Z}$  &  deconv 1   &  $1\times3\times3\times30$   &  (1,0) \\
			 &     deconv 2  &   deconv 1   &  deconv 2   &   $1\times3\times3\times20$  &  (2,1) \\			
			 &     deconv 3  &   deconv 2   &  $\mathbf{\hat X}$   &   $1\times5\times5\times10$  &  (2,1) \\	
			\hline			
		\end{tabular} 
	}
\end{table}

\noindent \textbf{Training details:}  We implemented our method with Tensorflow-1.4 \cite{abadi2016tensorflow}. We use the adaptive momentum-based gradient descent method (ADAM) \cite{kingma2014adam} to minimize the loss function, and apply a learning rate of $10^{-3}$.  Before we start training on our objective function, in each experiment, we pre-train our encoder and decoder on the dataset without the sparse coding layer.		  In particular, we pre-train the encoder-decoder for $20k$ epochs with the objective of
$\min_{\hat{\boldsymbol{\Theta}}}  \| \mathbf{X} -  \hat{\mathbf{X}}\|^2_F,$ 
where $\hat{\boldsymbol{\Theta}}$ indicates the union of parameters in the encoder and decoder networks.	We use a batch size of $100$ for this stage. However, in the actual stage of training, we feed all the samples including the training and testing samples as a single large batch.  We set the regularization parameters as $\lambda_0 =1$ and $\lambda_1= 8$ in all the experiments. 

\begin{table*}[t]
\begin{center}
\resizebox{\linewidth}{!}{%
\begin{tabular}{|l|c|c|c|c|c|c|c|c|c|}
\hline
 Dataset  & 	SRC	 &  	KSRC   &  AE-SRC  &  VGG19-SRC &  InceptionV3-SRC &  Resnet50-SRC &  Denesnet169-SRC  &  DSRC (ours) \\
\hline\hline
USPS  & 	87.78  &  91.34  &  88.65 &  91.27 & 93.51 &  95.75 &  95.26 &  \textbf{96.25}	 \\
\hline
SVHN  & 	15.71  &  27.42 &  18.69  & 52.86 & 41.14 &  47.88 &  37.65 &  \textbf{67.75}	 \\
\hline
UMDAA-01  & 	79.00  &  81.37  &  86.70 &  82.68  & 86.15 &  91.84 &  86.35 &  \textbf{93.39}	 \\
\hline
\end{tabular}}
\caption{Sparse representation-based classification accuracy of different methods. } \label{tbl:digits}
\end{center}
\vskip -22pt
\end{table*}

\subsection{USPS digits}
The first set of experiments is conducted on the USPS handwritten digits dataset~\cite{hull1994database}.  This dataset contains 7291 training and 2007 test grayscale images of ten digits (0-9).  Figure~\ref{fig:datasets} (a), shows example images from this dataset.   We perform the experiments on a  subset with a total size of $2000$ samples. In particular, we randomly select $160$ and $40$ samples per digit from the training and testing sets, respectively.  The first row of Table~\ref{tbl:digits} shows the performance of various SRC methods.  As can be observed from this table, the proposed method performs significantly better than the other methods including the classical and deep learning-based methods.

Figure~\ref{fig:res} shows the coefficient matrix $\mathbf{A}$, extracted from $\boldsymbol{\Theta_{sc}}$, the matrix of the network trained for this experiment.  For better visualization, we show the absolute value of the transposed $\mathbf{A}$~(i.e. $|\mathbf{A}^T|$).  Thus, each row of the matrix in Figure~\ref{fig:res} corresponds to the sparse codes for one of the test samples. Similarly, columns in this figure are coefficients related to the training samples. This matrix is sparse and shows a block diagram pattern, where most of the non-zero coefficients for each test sample are those that correspond to the training samples with the same class as the observed test sample. 

\noindent \textbf{Analysis of the network:} To understand the effects of some of our model choices, we compare the performance of our DSRC method with variations of it by changing the regularization norm on $\boldsymbol{\Theta_{sc}}$ in the loss function \eqref{eq:loss}.  We replace the term $ \|\boldsymbol{\Theta}_{sc} \|_{1}$ in \eqref{eq:loss} by $\|\boldsymbol{\Theta}_{sc} \|_{p}$, where $p=0.5,1.5$ and $2$, and report their performances by \emph{DSRC$_{0.5}$, DSRC$_{1.5}$} and \emph{DSRC$_{2}$}, respectively.  

In addition, if we do not follow the specific structure described in equation~\eqref{eq:theta}, and instead have a fully connected layer with $(m+n)^2$ parameters which receives $\mathbf{Z}$ and reconstructs  $\mathbf{\hat Z}$, the architecture of the network will be similar to the deep subspace clustering networks (DSC) proposed in~\cite{deepsc17nips} for the task of subspace clustering. As an ablation study, we use this method to extract sparse codes and then apply the same classification rule as in~\eqref{eq:classify} to estimate class labels for the test set. We call this method \emph{DSC-SRC}.

Table~\ref{tbl:ablation} reveals that while the regularization norm on the coefficient matrix is selected between $\ell_1$ and $\ell_2$, it does not have much effect on the performance of the classification task. However, in our experiments, we observed that for norms smaller than 1, the problem is not stable and often does not converges.  In addition, \emph{DSC-SRC} cannot provide a desirable performance.  Note that the fully-connected layer in this method (counterpart to our sparse coding layer) does not limit the testing features to be reconstructed with only the training features. As a result,  it is possible that testing features shape an isolated group that does not have a strong connection to the training features. This makes it more difficult to estimate a label for the test samples. 

\begin{table}[t]
\begin{center}
\resizebox{\linewidth}{!}{%
\begin{tabular}{|l|c|c|c|c|c|}
\hline
  & 	DSRC	 &  	DSC-SRC    &  DSRC$_{0.5}$  &  DSRC$_{1.5}$  & DSRC$_{2}$\\
\hline\hline
USPS  & 	\textbf{96.25}  &  78.25 &   \texttt{\small{N/C}}   &  95.75  &  \textbf{96.25}	 \\
\hline
\end{tabular}}
\caption{The classification accuracy corresponding to the ablation study.  \texttt{\small{N/C}} refers to the cases where the learning process did not converge. }
 \label{tbl:ablation}
\end{center}
\vskip -13pt

\end{table}

\vskip -30pt
\subsection{Street view house numbers}
The SVHN dataset~\cite{netzer2011reading} contains  630,420 color images of real-world house numbers collected from Google Street View images.  This dataset has three splits as the training set with 73,257 images, the testing set with 26,032 images and an extra set containing 531,131 additional samples. In this experiment, similar to our experiments on MNIST,  we randomly select 160 images per digit from the training split and 40  per digit from the test split.  This dataset is much more challenging than MNIST. This is in part due to the large variations of data.  Furthermore, many samples in this dataset contain multiple digits in an image.  The task is to classify the center digit. 

The second row in Table~\ref{tbl:digits} compares the performance of different SRC methods.  This table demonstrates the advantage of our method.  While the classification task is much more challenging on SVHN than MNIST, the gap between the performance of our method and the second best performance is even more.  The next best performing method is \emph{VGG19-SRC} which  performs $14.86\%$ behind the accuracy of our method.

\subsection{UMD mobile faces}
The UMD mobile face dataset (UMDAA-01)~\cite{zhang2015domain} contains 750 front-facing camera videos of 50 users captured while using a smartphone.  This dataset has been collected over three different sessions.  This dataset was originally collected for the active authentication task, but since its frames include challenging facial image instances with various illumination and pose conditions it has also been used for other tasks~\cite{abavisani2016domain,abavisani2018adversarial}.  In this experiment, we randomly select 50 facial images per subjects from the data in Session 1. Figure \ref{fig:datasets} shows some sample images from this dataset. 

The performance of various SRC methods on the UMDAA-01 dataset are tabulated in the third row of Table~\ref{tbl:digits}. As can be seen, our proposed DSRC method similar to the experiments with SVHN provides remarkable improvements as compared to the other SRC methods.  This clearly shows that more challenging datasets are better represented by our method.   This is because our method not only efficiently finds the sparse codes, but also it seeks for a representation of data (the output of the encoder) that is especially suitable for sparse representation.

\noindent \textbf{Comparison to state-of-the-art classification networks:}
While deep neural networks perform very well when they are trained on large datasets, in the case of limited number of labeled training samples, they often tend to overfit to the training samples.  The objective of this experiment is to analyze the performance of our approach in such circumstances.   We compare our method to the following classification networks: VGG-19\cite{simonyan2014very}, Inception-V3~\cite{szegedy2016rethinking}, Resnet-50~\cite{he2016deep} and Densenet-169~\cite{huang2017densely}.   We first pre-train the networks on the Imagenet dataset~\cite{deng2009imagenet}, and then fine-tune them on the available training samples in UMDAA-01. 

\begin{figure}[t]
 \centerline{\begin{overpic}[width=0.5\textwidth,tics=3]{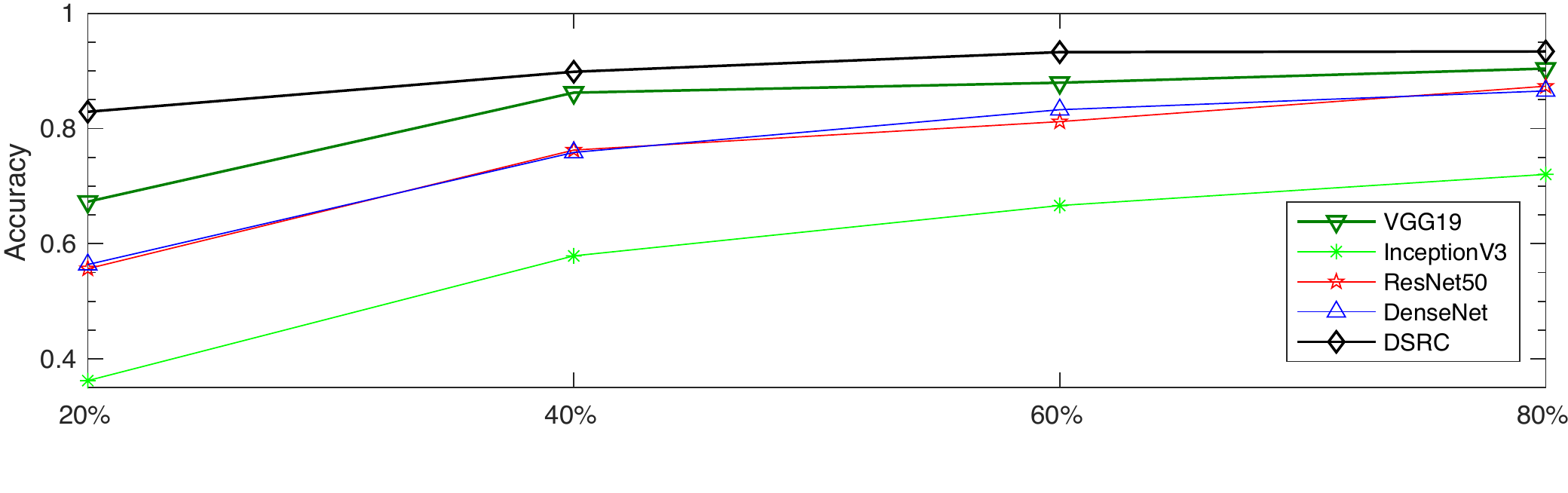}
 \put (41,2) {\tiny{$\frac{\text{\# training samples}}{\text{total \# samples}}\times 100$}}
 \end{overpic}}
\vskip -12pt\caption{Effect of the number of training samples on the performance of different classification networks.  The figure shows five-fold averaged classification accuracies of the methods trained on varying number of training samples in the UMDAA-01 dataset.}
\label{fig:lowlabelrates}
\end{figure}

Figure~\ref{fig:lowlabelrates} shows the performance of the classification networks on four different versions of UMDAA-01 dataset with varying number of training samples.   The four versions are created by randomly splitting the dataset into sets of training and testing samples that respectively contain $20\%$, $40\%$, $60\%$ and $80\%$ of the total number of samples as training samples and use the rest of samples as the testing set.   As the figure suggests, accuracy improves by increasing the number of training samples in all the cases. However, the results show better performances for DSRC even when less training data is available.

\section{Conclusion}\label{sec:conclusion}
We presented an autoencoder-based sparse coding network for SRC.  In particular, we introduced a sparse coding layer that is located between the conventional encoder and decoder networks.   This layer recovers sparse codes of embedding features that are received from the encoder. The spare codes are later used to estimate the class labels of testing samples.  We discussed a framework that allows an end-to-end training.  Various experiments on three different image classification datasets showed that the proposed network leads to sparse representations that give better classification results than state-of-the-art SRC methods. 

\bibliographystyle{IEEEbib}
\bibliography{DSRC_CR}
\newpage 
 \section*{Appendix: Convergence}
\begin{figure}[h]
	\centerline{\includegraphics[width=.55\textwidth]{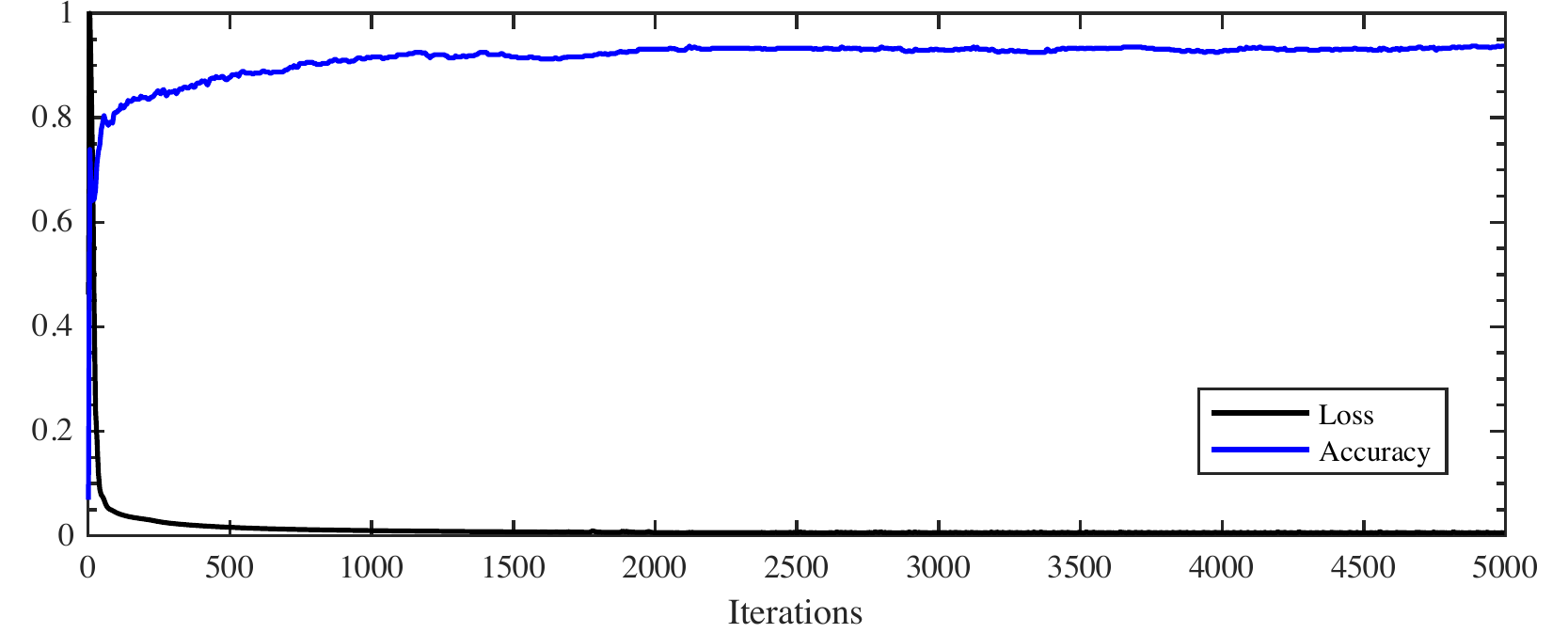}}
	\caption{Values of the DSRC's loss function in the experiment on the UMDAA-01 dataset vs iterations. The loss values are scaled to have the maximum value of one.}
	\label{fig:convergence}
\end{figure}

To empirically show the convergence of our proposed method, in Figure~\ref{fig:convergence}, we plot the values of the objective function of DSRC in the experiment with the UMDAA-01 dataset and its classification accuracy in different iterations.  The reported loss values in Figure~\ref{fig:convergence} are scaled to have a maximum value of one.   As can be seen from the figure, our algorithm converges in a few iterations.

\end{document}